\documentclass[a4paper,conference]{IEEEtran}

\ifCLASSINFOpdf
\else
\fi
\usepackage{times}
\usepackage{epsfig}
\usepackage{graphicx}
\usepackage{amsmath}
\usepackage{amssymb}
\usepackage{url}
\usepackage[hidelinks]{hyperref}
\usepackage[utf8]{inputenc}
\usepackage[small]{caption}
\usepackage{graphicx}
\usepackage{amsmath}
\usepackage{booktabs}
\usepackage{algorithm}
\usepackage{algorithmic}
\usepackage{times}
\usepackage{epsfig}
\usepackage{graphicx}
\usepackage{amsmath}
\usepackage{amssymb}
\usepackage{array}
\usepackage{enumitem}
\newcolumntype{L}[1]{>{\raggedright\let\newline\\\arraybackslash\hspace{0pt}}m{#1}}
\newcolumntype{R}[1]{>{\raggedleft\let\newline\\\arraybackslash\hspace{0pt}}m{#1}}
\newcolumntype{C}[1]{>{\centering\let\newline\\\arraybackslash\hspace{0pt}}m{#1}}
\begin{document}

\title{Age Gap Reducer-GAN for Recognizing Age-Separated Faces}

\begingroup
\centering{
\author{Daksha Yadav$^1$, Naman Kohli$^1$, Mayank Vatsa$^2$, Richa Singh$^2$,  Afzel Noore$^3$\\
$^1$West Virginia University, $^2$IIT Jodhpur, $^3$Texas A\&M University-Kingsville \\
{\tt\small $^1$\{dayadav, nakohli\}@mix.wvu.edu,$^2$\{mvatsa, richa\}@iitj.ac.in}, $^3${\tt\small afzel.noore@tamuk.edu} }}
 \endgroup

\maketitle

\begin{abstract}
 In this paper, we propose a novel algorithm for matching faces with temporal variations caused due to age progression. The proposed generative adversarial network algorithm is a unified framework that combines facial age estimation and age-separated face verification. The key idea of this approach is to learn the age variations across time by conditioning the input image on the subject's gender and the target age group to which the face needs to be progressed. The loss function accounts for reducing the age gap between the original image and generated face image as well as preserving the identity. Both visual fidelity and quantitative evaluations demonstrate the efficacy of the proposed architecture on different facial age databases for age-separated face recognition.
\end{abstract}

\IEEEpeerreviewmaketitle

\section{Introduction}
The research in face recognition has witnessed a significant increase in the performance due to the development of different deep learning models \cite{Parkhi15,schroff2015facenet,wu2018light, taigman2014deepface}
. While building these algorithms, the emphasis is on developing an algorithm that is robust to variations such as pose, illumination, expression, disguise, and makeup \cite{le2019illumination,masi2018learning, rsmv2}. A critical challenge of face recognition is matching face images with temporal variations, specifically age gaps \cite{du2019cycle}. Building age-invariant face recognition algorithms can prove to be beneficial in many applications such as locating missing persons, homeland security, and passport services. In fact, for large-scale applications, adding invariance to aging is a very important requirement for face recognition algorithms.

During the lifetime of an individual, temporal variations alter the facial appearance in diverse ways. Different studies have reported that every person has a personalized aging pattern depending on numerous factors including ethnicity, environmental conditions, and stress level \cite{rsmv3}, \cite{Farkas, ChinaWomen}. Apart from these factors which contribute to the complexity of facial aging, limited availability of labeled databases makes the problem of matching age-separated faces, extremely challenging.

In the literature, various approaches have been proposed for matching faces with age progression and are mainly categorized as generative or discriminative \cite{jainaging}. Discriminative methods involve finding the age-invariant signatures from the input faces and use it for the recognition task. Generative methods involve inducing the changes in the input facial images to incorporate aging variations and projecting the images at a common age which is followed by similarity computation of the two faces. Recently, generative adversarial networks (GANs) \cite{goodfellow2014generative} are being utilized to generate synthetic images using convolutional neural networks. Due to their popularity, different GAN based approaches have been proposed for facial age simulation \cite{liu2019attribute,Yang_2018_CVPR}.

\begin{figure}[t]
	\centering
	\includegraphics[width=1.\linewidth]{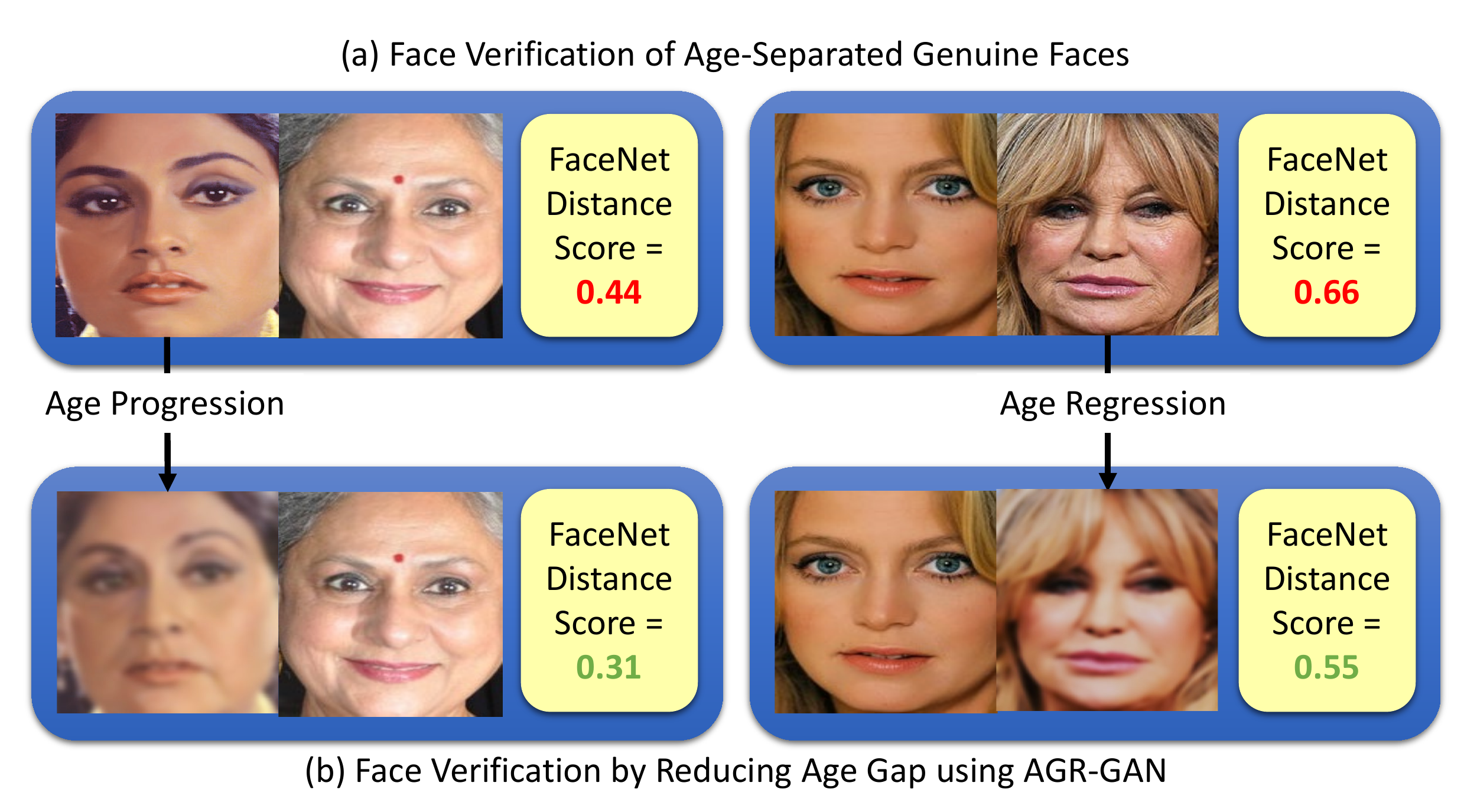}
	\caption{Illustrating the problem of age-separated face recognition and the solution proposed in this paper using Age Gap Reducer-Generative Adversarial Network (AGR-GAN). The proposed AGR-GAN is utilized in conjunction with an existing face recognition model such as FaceNet \cite{schroff2015facenet} to decrease the distance score of age-separated faces (best viewed in color).}
	\label{fig:IntroFig}
\end{figure}

The majority of existing GANs based research related to facial aging focus only on generating images for different age groups. Moreover, only some of these techniques can produce both age-progressed as well as age-regressed faces and very few of them cater to both young as well as old age groups. Most of these techniques do not demonstrate their efficacy in enhancing the face recognition accuracy of age-separated probe and gallery face images. In this paper, we propose Age Gap Reducer-Generative Adversarial Network (AGR-GAN) where the focus is on reducing the age gap between face images by using automatic age estimation. The input image is conditioned on the individual's gender as well as the target age group to which the input face needs to be progressed. The key advantage of this approach is that an input image can be regressed to an earlier age group or progressed to an older age group. We demonstrate the efficacy of the proposed AGR-GAN in enhancing the performance of face recognition algorithms in matching age-separated probe and gallery images.

\begin{figure*}[t]
	\centering
	\includegraphics[width=.7\linewidth]{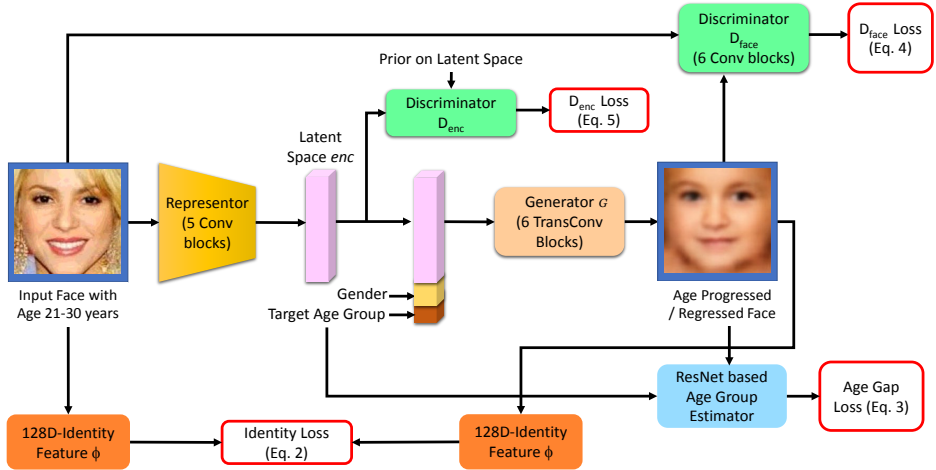}
	\vspace{-1.5cm}
	\caption{Proposed Age Gap Reducer-Generative Adversarial Network (AGR-GAN) architecture.}
	
	\label{fig:GANArch}
\end{figure*}

\subsection{Related Work}
Lanitis et al. \cite{Lanitis2004, Lanitis2002} proposed utilizing the training images for finding the relationship between the coded face representation and the facial age of the subject. This relationship is then utilized for estimating the age of a facial image and simulating the facial appearance at any given age. Singh et al. \cite{rsmv1} proposed image registration in polar domain to minimize the variations in facial features caused due to aging. Park et al. \cite{Park} developed a 3D facial aging model to address the problem of age-invariant face recognition. Their approach is based on the fact that exact craniofacial aging can be developed only in the 3D domain. Li et al. \cite{jainaging} proposed a discriminative model for age-invariant recognition. They developed an approach involving the use of scale invariant feature transform (SIFT), multi-scale local binary pattern as local descriptors, and multi-feature discriminant analysis.

In the recent literature, different GAN based generative approaches have been proposed for facial age simulation. For instance, Wang et al. \cite{wang2016recurrent} generated faces across different ages by learning the intermediate transition states using a recurrent neural network. However, this approach requires multiple face images at various ages per subject. Duong et al. \cite{duong2017temporal} developed different approaches for the short-term age progression and long-term age progression. However, it did not consider any personality/identity information which resulted in varying identity in the generated faces. Yang et al. \cite{Yang_2018_CVPR} proposed a GAN based approach to progress face images with $<$ 30 age to older age groups. Such GAN based approaches can be utilized for reducing the age gap in age-separated face images for performing face recognition. Liu et al. \cite{liu2019attribute} utilized wavelet-based GANs to perform attribute-aware facial aging. More recently, Duong et al. \cite{duong2019automatic} demonstrated face aging in videos using deep reinforcement learning. Jia et al. \cite{jia2018face} developed a face aging network using invertible convertible GAN without any data pre-processing steps. Genovese et al. \cite{genovese2019towards} presented an approach for increasing the explainability of current GAN models by analyzing common filters in different CNNs. Fang et al. \cite{fang2020triple} used triple translation loss to model the variations in aging patterns among different age groups. Zhu et al. \cite{zhu2020look} utilized attention mechanism to only alert the regions relevant to face aging.

In this research, we propose a unified solution which incorporates facial age estimation and age-separated face verification using GANs. As shown in Fig. \ref{fig:GANArch}, we propose Age Gap Reducer-Generative Adversarial Network (AGR-GAN) for this task. In the proposed AGR-GAN, the emphasis is on reducing the age gap between face images by using automatic age estimation and incorporating it in the overall loss function. The input image is conditioned on the individual's gender as well as the target age group to which the input face needs to be progressed. Additionally, the training critic simultaneously learns the age group of an input image while estimating how realistic the face appears. An additional constraint is placed on the loss function of AGR-GAN to account for keeping consistent identity across age progression by embedding the generated face closer towards the input image in a lower dimensional face subspace. The key advantage of this approach is that the age synthesis is bi-directional and given an input image, it can be regressed to an earlier age group or progressed to an older age group (as demonstrated in Fig. \ref{fig:IntroFig}). The contributions of this research are as follows:
\begin{itemize}
	\item A novel approach for matching age-separated face images is proposed. To accomplish this, a novel GAN architecture, AGR-GAN, is proposed which uses a multi-task discriminator that is able to progress/regress the age of an input face to a target age group. Apart from the traditional GAN loss, the proposed AGR-GAN incorporates an identity preserving feature which ensures that the generated (regressed/progressed) face image has the same identity representation as the input face image.  
	\item In the proposed AGR-GAN, joint learning of the age group estimator module with the image generation is performed. This novel architecture eliminates the need for age-labeled data in the training phase.
	\item The efficacy of the proposed AGR-GAN is demonstrated on three publicly available facial aging databases for the problem of age-separated face recognition.
\end{itemize}

\section{Proposed Age Gap Reducer-GAN}

Generative Adversarial Network consists of a generator network ($G$) and a discriminator network ($D$) which competes in a two-player minimax game \cite{goodfellow2014generative}. The aim of $D$ is to distinguish real images from fake images generated by $G$ and $G$ aims to synthesize \textit{real-looking} images to fool $D$. Specifically, $D$ and $G$ play the minimax game with a value function $V$ such that:

\begin{equation}
\begin{aligned}
\underset{G}{min} \: \underset{D}{max} \: V = & E_{x\sim P_{d}(x)} \: [\log D(x)] \\ +  
& E_{z\sim P_{z}(z)} \: [\log (1-D(G(z)))]
\end{aligned}
\end{equation}

\noindent where $z$ is a noise sample from a prior probability distribution $P_{z}$ and input face $x$ follows a specific data distribution $P_{data}$. Upon convergence of the network, the distribution of generated faces $P_{g}$ should become equivalent to $P_{data}$. Traditionally, $G$ and $D$ networks are trained alternatively.

In the proposed AGR-GAN, the input and output face images are $128 \times 128$ RGB images. The input face image is encoded through a representor network to form a low dimension representation $enc$ which learns the higher level features of the input face image. Using the learned $enc$ and conditional information of gender ($g$) as well as target age group ($a$), the generator network generates the output face image ($x'$). We apply an adversarial loss on $enc$ ($D_{enc}$) to ensure it is uniformly distributed \cite{zhang2017age}, thus, leading to smooth age transformations. Additionally, an adversarial loss, $D_{face}$, is utilized to ensure that the generated images are realistic looking. Using the target age group $a$ and the estimated age group of the generated face image, the age gap reduction loss is computed to minimize the age gap. Lastly, the feature representations of the generated face and input face image are computed using face feature mapping ($\phi$) to encode identity representations and preserve the identity information after age progression/regression. 

\subsection{Components of AGR-GAN}
Next, we provide in-depth details of the different components of the proposed AGR-GAN.

\begin{itemize}
    \item \textbf{Representor $R$}: An input face image is passed through the representor network $R$ whose aim is to learn its low-dimensional representation $enc$. The input RGB face image $x$ of size $128 \times 128$ is sent to the representor network $R$ consisting of five blocks of convolutional layers with stride = 2 and $5 \times 5$ convolutional kernels followed by an exponential linear unit layer \cite{ClevertUH15} to learn the facial features which are invariant to age progression/regression. Each convolutional layer is followed by spectral normalization which aids in stabilizing the training phase \cite{miyato2018spectral}. After the convolutional layers, a fully connected layer is applied to compute the low-dimensional encoding $enc$ ($R(x) = enc$).
    \item \textbf{Generator $G$}: The objective of the generator network $G$ is to utilize $enc$ to synthesize a face image $x'$ with the same gender $g$, and target age group $a$. $enc$ is concatenated with the gender label ($g$) and expected age group label ($a$). This feature vector is now processed by the decoding part of the generator which consists of a fully connected layer followed by six blocks of transposed convolutional layers with stride = 2 and padding = 2. Each deconvolutional layer is followed by an Exponential Linear Unit (ELU) layer except the last layer which is succeeded by a $tanh$ layer. The output face from the generator is of size $128 \times 128 \times 3$.
    \item \textbf{Discriminator $D_{face}$}: Similar to traditional GANs, the objective of discriminator $D_{face}$ is to distinguish synthetically generated images by $G$ from real images. It consists of six blocks of convolutional layers with kernel size = 5, stride = 2, and padding = 2. Each convolutional layer is followed by the ELU layer. Each convolutional layer is also followed by spectral normalization which aids in stabilizing the training phase \cite{miyato2018spectral}. Lastly, a sigmoid layer is utilized to classify the image as real/fake.
    \item \textbf{Discriminator $D_{enc}$}: The discriminator on $enc$ ($D_{enc}$) ensures that the distribution of the encoded feature vector $enc$ approaches the prior distribution. The goal of $D_{enc}$ is to distinguish $enc$ generated by $R$ as compared to the prior distribution. On the other hand, $R$ is forced to generate $enc$ such that it can fool $D_{enc}$. This ensures that $enc$ smoothly populates the low-dimensional latent space \cite{zhang2017age} to remove unrealistic faces. Our experimental analysis revealed that using this adversarial loss resulted in better images as compared to KL loss.
    \item \textbf{Age Group Estimator}: To reduce the age gap between the input face ($x$) and the generated face image ($x'$), an age group estimator module is used in the proposed formulation. Given an input image, the age group estimator utilizes ResNet-18 model \cite{he2016} as its backbone to predict the age group of the input image.  For this purpose, a pretrained ResNet-18 model is finetuned to predict the correct age group. An adaptive average (spatial) pooling layer is utilized for removing the limitation of the fixed size of the input image and a fully connected layer with output size 10 is employed to predict the final age group. The loss function utilized to train this network is the sum of cross-entropy loss between the correct and predicted age group and mean average group error.
\end{itemize}

\subsection{Objective Function}

Given an input face image $x$, the representor network $R$ maps it to a latent space to learn the encoding $enc$. Given this learned $enc$, age group $a$, and gender $g$, the generator network $G$ synthesizes a face image $x'$. To ensure that the identity of the generated face image ($x'$) is same as the input face image, we compute the identity-specific feature mapping using the function $\phi$ and minimize the cosine distance $CosDist$ between these two mappings. In this formulation, Light-CNN \cite{wu2018light} is used as the identity-specific feature mapping ($\phi$) which has been trained to generate similar feature representations for the same identity. Thus, the identity loss (ID loss) for this component is:
\begin{equation}
\begin{aligned}
IDLoss = & \underset{G,R}{min} \:  CosDist(\phi(x'), \phi(x)) \\ 
     = & \underset{G,R}{min} \: CosDist( \phi(G(R(x),a,g)), \phi(x))
\end{aligned}
\end{equation}

Simultaneously, the age group estimator in $G$ is trained by comparing its output with the ground truth age group label using $L_{1}$ loss (age gap loss). 
\begin{equation}
Age Gap Loss = \underset{}{min} \: L_{1}(a, a')
\end{equation}

With respect to the face image based discriminator $D_{face}$ which is conditioned on the age group $a$ and gender label $g$, it is trained by the following loss function:
	
\begin{equation}
\begin{aligned}
& D_{face}Loss = \\
& \underset{G}{min} \:  \underset{D_{face}}{max} \: 
E_{x,a,g\sim p_{train}(x,a,g)} \: [\log D_{face}(x, a, g)]  + \\
& E_{x,a,g\sim p_{train}(x,a,g)} \: [\log (1-D_{face}(G(R(x), a, g)))]
\end{aligned}
\end{equation}

Likewise, the discriminator on the encoded feature representation ($enc$) from the Representator network $R$ is trained to ensure that $enc$ follows the prior distribution using min-max objective function:

\begin{equation}
\begin{aligned}
D_{enc}Loss = 
& \underset{R}{min} \: \underset{D_{enc}}{max} \: E_{enc^{*}\sim p(enc)} \: [\log D_{enc}(enc^{*})] \: +  \\
& E_{x\sim p_{train}(x)} \: [\log (1-D_{enc}(R(x))] 
\end{aligned}
\end{equation}

To minimize perceptual loss and remove any ghosting artifacts, total variation loss is computed as follows:
\begin{equation}
TVLoss = \underset{G, R}{min} \; TV(G(R(x), a, g))
\end{equation}

By combining the different loss functions, the overall objective function becomes:
\vspace{-5mm}

\begin{equation}
\begin{aligned}
& L_{AGR-GAN}=\\
& \underset{G, R}{min} \; \underset{D_{enc}, D_{face}}{max} \: CosDist( \phi(G(R(x),a,g)), \phi(x)) \\
& + L_{1} (a, a') + TV(G(R(x), a, g)) \\
& +  E_{x,a,g\sim p_{train}(x,a,g)} \: [\log (1-D_{face}(G(R(x), a, g)))] \\
& + E_{x,a,g\sim p_{train}(x,a,g)} \: [\log D_{face}(x, a, g)] \\
& + E_{x\sim p_{train}(x)} \: [\log (1-D_{enc}(R(x)))] \\
& + E_{enc^{*}\sim p(enc)} \: [\log D_{enc}(enc^{*})] 
\end{aligned} 
\end{equation}

\subsection{Implementation Details}
Before providing the images as input to the network, the following pre-processing steps are performed. Face detection and alignment are performed using pretrained MTCNN model \cite{mtcnn} which produces an output face image of size $128 \times 128 \times 3$. Next, age labels are arranged in bins to create age groups. During the initial years, significant variations are observed in the facial appearance; hence the bins are of size 5. Starting from the age of 20 years, bins (20-30 years, 30-40 years, \ldots, \textgreater=60 years) are created of size 10. Next, the age group labels and the gender labels are transformed to one-hot encoding. Image normalization is performed to scale the intensity values to the range of [-1, 1]. A batch size of 128 is utilized. Adam optimizer with a learning rate = 0.0002 and momentum = 0.5 is used. For training $D_{enc}$, prior of uniform distribution is utilized. 

\section{Experimental Evaluation}
To demonstrate the efficacy of the proposed AGR-GAN, five experiments are conducted (i) visual fidelity, (ii) aging model evaluation, (iii) identity preservation across generated faces, (iv) ablation study, and (v) age-separated face recognition. 

\begin{figure*}[!t]
	\centering
	\includegraphics[width=.85\linewidth]{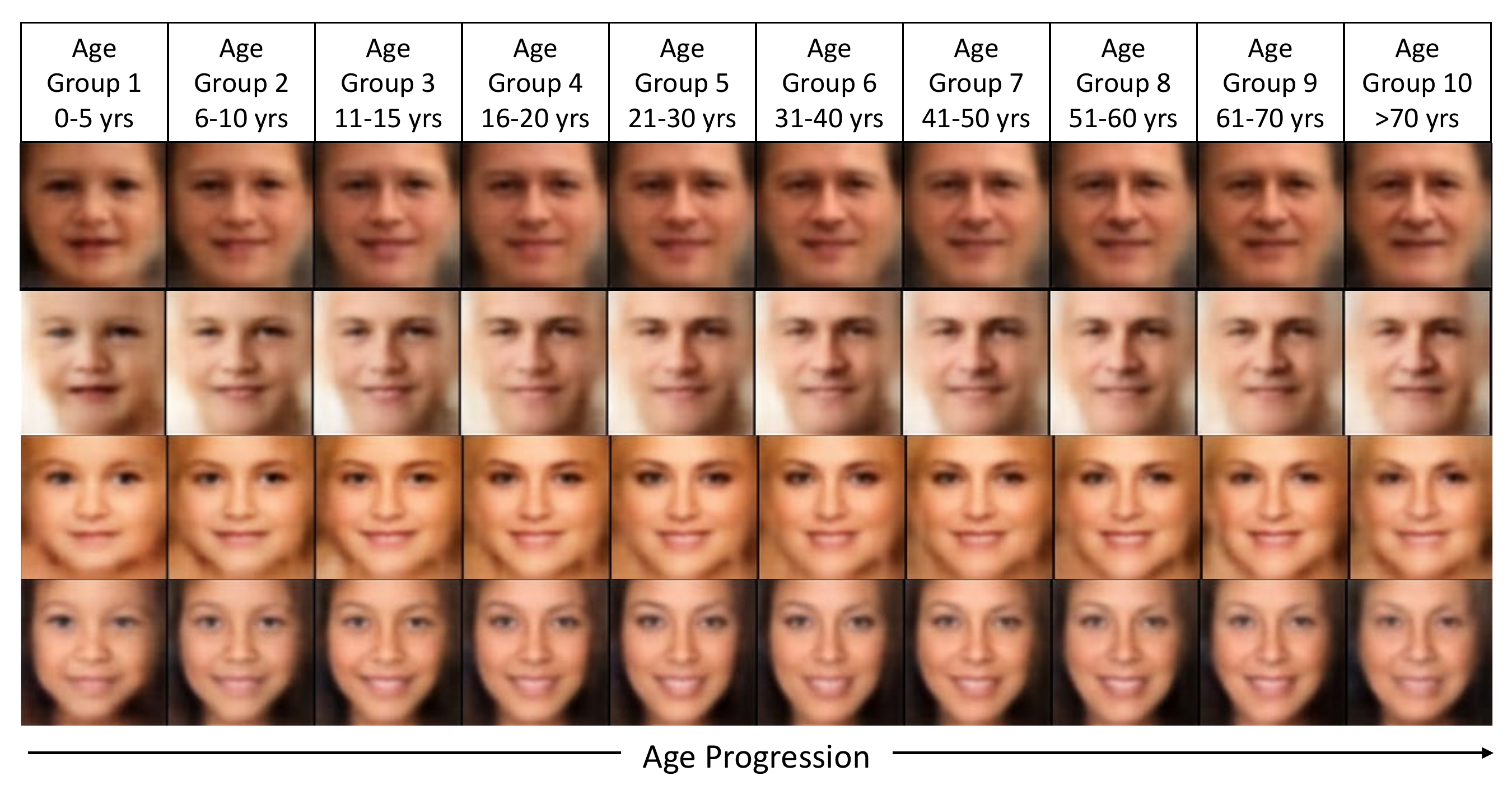}
	
	\caption{Sample generated outputs by the proposed Age Gap Reducer-GAN across different age groups. A single row contains outputs for the same subject across the different age groups. Subjects change in each row. Subjects taken from UTKFace \cite{zhang2017age} and CALFW \cite{calfw}.}
	\label{fig:GanSamples}
\end{figure*}

\subsection{Databases}
For evaluation, we utilized MORPH \cite{morph}, CALFW \cite{calfw}, UTKFace \cite{zhang2017age}, and CACD-VS \cite{chen2015face} datasets.
\begin{itemize}
    \item MORPH database comprises more than 55,000 age-separated face images of 13,000 subjects. The age range of face images in this database is 16-70 years with the average age = 33 years. Similar to the protocol described in \cite{chen2015face}, the training set contains face images from 10,000 subjects. The test set is formed using probe and gallery sets from the remaining 3000 subjects and each subject has two images with the largest age gap.
    \item CALFW database \cite{calfw} is built on the concept of the popular LFW face database but with age gaps to add aging as another factor contributing to intra-class variance. It includes 3000 positive face pairs with age gap and an equal number of negative pairs with 10-fold cross-validation sets.
    \item UTKFace dataset \cite{zhang2017age} is a large-scale database with more than 25,000 images with a long age span of 0 to 116 years. It does not contain labeled identity information and therefore, is utilized for training purposes only. 
    \item  The CACD database released by Chen et al. \cite{chen2015face} consists of 160,000 images of 2000 celebrities. Even though the database is very challenging due to wild variations including age, it contains noisy labels. To counteract this, Chen et al. \cite{chen2015face} released the CACD-VS subset which contains 2000 positive pairs and 2000 negative pairs which are manually checked. For evaluation, the CACD-VS database is split into 10 cross-validation folds with 200 positive pairs and 200 negative pairs in the testing.
\end{itemize}

For training the proposed AGR-GAN, the training folds of MORPH, CACD-VS, and CALFW, and the complete UTKFace are combined to creating the training set. Testing is performed on the test sets of MORPH, CACD-VS, and CALFW databases.

\subsection{Experimental Analysis}
In this section, we describe the different experimental settings and quantitative analysis for evaluating the face images generated by the proposed AGR-GAN architecture.

\renewcommand{\arraystretch}{1.0}
\begin{table*}[!t]
\centering
\caption{Increase in FaceNet model based face recognition performance by using faces generated from the proposed AGR-GAN on: (a) MORPH, (b) CACD-VS, and (c) CALFW. The column `Per-DB SOTA' refers to the state-of-the-art performance reported on the databases using the protocol mentioned in Section III.A.}

\begin{tabular}{L{2.5cm}L{1.75cm}C{2.cm}C{2.5cm}C{1.7cm} C{2.4cm}}
\toprule
\textbf{DB} & \textbf{Metric}& \textbf{Per-DB SOTA} & \textbf{Only FaceNet}   & \textbf{FaceNet + IPCGAN \cite{wang2018face}} & \textbf{FaceNet + AGR-GAN}\\ \midrule

(a) MORPH       & Rank-1  & 93.60 \cite{li2018distance}         		  &    94.03     & 94.10  &  \textbf{94.15} \\ \hline

(b) CACD-VS      & Accuracy @ FPR=0.1\% & 91.10 \cite{li2018distance}   &    97.50    & 98.04 &  \textbf{98.39}   \\ \hline

(c) CALFW      & Accuracy @ FPR=0.1\%  & 86.50 \cite{calfw}   &    57.50    & 85.23 &  \textbf{87.15}   \\ \bottomrule

\end{tabular}
\label{tab:verification}
\end{table*}

\subsubsection{Visual Fidelity of Aging Simulation}
To evaluate the efficacy of AGR-GAN in synthesizing images across the different age groups, the testing sets of the databases are utilized. Fig. \ref{fig:GanSamples} demonstrates the synthesis output of multiple subjects from different databases across the 10 age groups. Upon visual analysis by different human evaluators, it is observed that the proposed GAN architecture is able to learn the aging patterns across different age groups as well as maintain the identity information across different synthesis outputs of the same subject. It is able to model the aging patterns even with varying facial hair, gender, and ethnicity. This illustrates that the proposed AGR-GAN is able to generate photo-realistic face images across different age groups. For some cases, the generated faces may appear over-smoothed. This may be attributed to the presence of $L_{1}$ term in the loss function which has been observed in other recent works as well \cite{8696010}. Another reason for this can be the insufficient number of training samples for different age groups, specifically the younger and older age groups.

\subsubsection{Age-Separated Face Recognition Accuracy}
The main objective of developing the age gap reducer GAN is to increase the performance of matching age-separated faces. To validate this, face recognition experiment is performed on the three databases. To demonstrate the efficacy of the proposed AGR-GAN in matching age-separated faces, we utilize FaceNet \cite{schroff2015facenet} as the baseline face recognition model. For the MORPH database, the test set containing the youngest image as a gallery and oldest image as a probe is used for performing the face identification experiment. For CACD-VS and CALFW databases, positive and negative face pairs are selected from each fold of cross-validation to perform the face verification experiment. Using the testing sets, we evaluate the performance of the FaceNet model and the results are reported in \textit{Only FaceNet} column of Table \ref{tab:verification}. 

\begin{figure}[!t]
	\centering
	\vspace{-20mm}
	\includegraphics[width=0.9\linewidth]{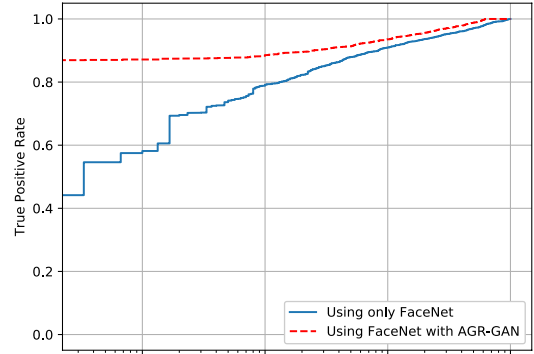}
	\vspace{-3cm}
	\caption{Receiver Operating Characteristic (ROC) curves demonstrating the increase in the performance of FaceNet by using AGR-GAN outputs on CALFW database. }
	\label{fig:roc}
\end{figure}

Next, AGR-GAN is applied on the input face pair and the query image is projected to the age group of the gallery face image. Using the age gap reduced face pair output from AGR-GAN, FaceNet model is re-evaluated. Table \ref{tab:verification} summarizes the result of this experiment. For all three databases, it is observed that utilizing AGR-GAN outputs with FaceNet increases the age-separated face matching performance. The highest increase is noticed for the CALFW database (also shown in Fig. \ref{fig:roc}) which contains real-world \textit{wild} variations in the images apart from the age gap. The increase in face verification performance is pronounced at a false positive rate (FPR) = 0.1\% where the accuracy increases by 29.65\% as compared to using only FaceNet.  
As mentioned in \cite{calfw}, VGG-face by \cite{Parkhi15} yields 86.50\% verification performance and Noisy Softmax by \cite{chen2017} yields 82.52\% verification performance at equal error rate on this database. On the other hand, using the outputs produced by the proposed AGR-GAN with FaceNet yields 92.62\% face verification accuracy at an equal error rate. These results highlight the challenging nature of the database as well as the increase in the performance of FaceNet after using AGR-GAN outputs.

A similar performance increase is also observed on the other databases. On the CACD-VS database, the face verification performance at FPR = 0.1\% increases by 0.89\% after using the outputs from the proposed AGR-GAN. The face identification on the MORPH database shows that the proposed AGR-GAN is able to enhance the rank-1 identification accuracy by 0.12\%. This improvement in the face recognition scores illustrates the efficacy of the proposed AGR-GAN in matching faces with age gaps. 

For comparative analysis, the performance of AGR-GAN is compared with a recently proposed GAN, IPCGAN \cite{wang2018face}, for facial aging. The results for the same are tabulated in Table \ref{tab:verification}. It can be observed that for all three databases, the increase in the face recognition accuracy due to AGR-GAN is higher as compared to the existing IPCGAN.

It should be noted that this AGR-GAN based framework can be combined with any existing face recognition model to boost the face recognition accuracy. To validate this, LightCNN \cite{wu2018light} is utilized as the baseline face recognition model. In this setting, employing AGR-GAN enhances the face recognition of LightCNN by 0.19\%, 0.08\%, and 1.12\% on MORPH, CACD-VS, and CALFW databases respectively.

\begin{table}[!t]

\centering
\caption{Age estimation (years) of faces generated by the proposed AGR-GAN on (a) MORPH, (b) CACD-VS, and (c) CALFW.}
\begin{tabular}{C{2cm}R{1.5cm}R{1.6cm}R{1.5cm}}
\toprule
\textbf{Age Group (Age Range)} & \textbf{MORPH} & \textbf{CACD-VS}  & \textbf{CALFW}\\ \midrule
1 \ \ \ (0-5)             &        5.26       &     8.45       &  6.79  \\ 
2 \ (6-10)                &      12.18        &     11.32      &   12.38  \\ 
3 (11-15)                 &     14.32         &     15.09      &  14.23  \\ 
4 (15-20)                 &     17.65         &     18.94      &  19.36  \\ 
5 (21-30)                 &     29.22         &     27.13      &  22.71  \\ 
6 (31-40)                 &     33.51         &     39.10      &  35.13  \\ 
7 (41-50)                 &     47.20         &     42.59      &  41.36  \\ 
8  (51-60)                &     54.19         &     53.72      &  58.75  \\ 
9 (61-70)                 &    63.69          &     68.24      &  63.84   \\ 
10 \  (\textgreater70)    &      69.85        &      74.32     &  78.38   \\ \bottomrule
\end{tabular}

\label{tab:agingAccu}
\end{table}

\subsubsection{Aging Model Evaluation}
Apart from analyzing the visual fidelity of the generated faces, it is critical to evaluate the ability of the proposed model to produce face images with the targeted age group. For this, we utilize Dex \cite{rothe2015dex}, an off-the-shelf age estimation model, and predict the age of every synthetically generated face image. The performance is evaluated by analyzing the mean estimated age for each age group and the results are shown in Table \ref{tab:agingAccu}. Ideally, the mean age values for the 10 age groups should be in the age range 0-5, 6-10, 11-15, 16-20, 21-30, 31-40, 41-50, 51-60, 61-70, and $>$ 70.

For the MORPH database, face images from 3000 subjects are used for testing. The mean estimated age of generated images for the 10 age groups is 5.26, 12.18, 14.32, 17.65, 29.22, 33.51, 47.20, 54.19, 63.69, and 69.85 years respectively. Apart from age group 1 (age range 0-5 years), age group 2 (age range 6-10 years), and age group 10 (age range $>$ 70 years), the mean age of GAN generated faces in all other age groups follows the expected trend. The divergence in the values of age groups 1, 2, and 10 may be attributed to less number of training face images in the training set. Similar trends are also observed for CACD-VS and CALFW databases. These results demonstrate the efficacy of the proposed AGR-GAN in generating face images of the targeted age groups.

\begin{table}[!t]
\centering
\caption{Equal error rate (\%) for face verification of input test faces and faces from the 10 target age groups generated by the proposed AGR-GAN on (a) MORPH, (b) CACD-VS, and (c) CALFW.}
\begin{tabular}{C{2cm}R{1.5cm}R{1.6cm}R{1.5cm}}
\toprule
\textbf{Age Group (Age Range)} & \textbf{MORPH} & \textbf{CACD-VS}  & \textbf{CALFW}\\ \midrule
1  \ \ \ (0-5)                &       12.97       &     8.23      &  4.72      \\ 
2 \ (5-10)                 &      9.82         &     8.94        &  3.37    \\ 
3  (11-15)                &    8.65        &      5.25      &  3.18    \\ 
4  (16-20)                &      6.94         &     1.27        &   2.94   \\ 
5   (21-30)               &       1.63        &     0.23       &  1.83   \\ 
6  (31-40)                &      0.48       &     0.12       &   1.36  \\ 
7 (41-50)                 &      1.04        &    0.03      &   0.34  \\ 
8  (51-60)                &     1.07       &    0.41      &   0.81  \\ 
9  (61-70)                &       2.38            &    0.83      &   0.74  \\ 
10 \ (\textgreater 70)                &     9.55         &    1.24       &   2.72    \\ \bottomrule
\end{tabular}

\label{tab:identityAccu}
\end{table}

\subsubsection{Identity Preservation across Generated Faces}
Another key aspect of the proposed AGR-GAN is to ensure that the identity/personality information of the subject is preserved across the generated face images. To evaluate this, FaceNet \cite{schroff2015facenet} model based face recognition framework is utilized. For each database, the input test image is matched with the corresponding generated image across the 10 age groups and the face verification score is computed. The face verification performance is showcased by calculating the equal error rate (EER) and the results are summarized in Table \ref{tab:identityAccu}. 

For all three databases, the face verification performance of input images with the generated faces is high, as indicated by the low equal error rates in Table \ref{tab:identityAccu}. This result substantiates that the proposed AGR-GAN is able to preserve the identity information in the generated face images and can be used for face matching purposes. 

For the MORPH database, the best face verification performance is observed when the input face image is matched with the face image generated for the age group 6 (31-40 years) with an equal error rate of 0.48\%. This indicates that the proposed AGR-GAN model is able to accurately learn the facial age characteristics of this specific age group. On the other hand, the highest EER of 12.97\% is observed for the age group 1 (0-5 years) which indicates the difficulty in matching very young faces. This can be attributed to drastic and diverse changes in the facial features during the childhood years, thus making it difficult to encode these varied manipulations. Similar results for this identity preservation experiment are also obtained on CACD-VS and CALFW databases.


\begin{figure}[!t]
	\centering
	\includegraphics[width=.8\linewidth]{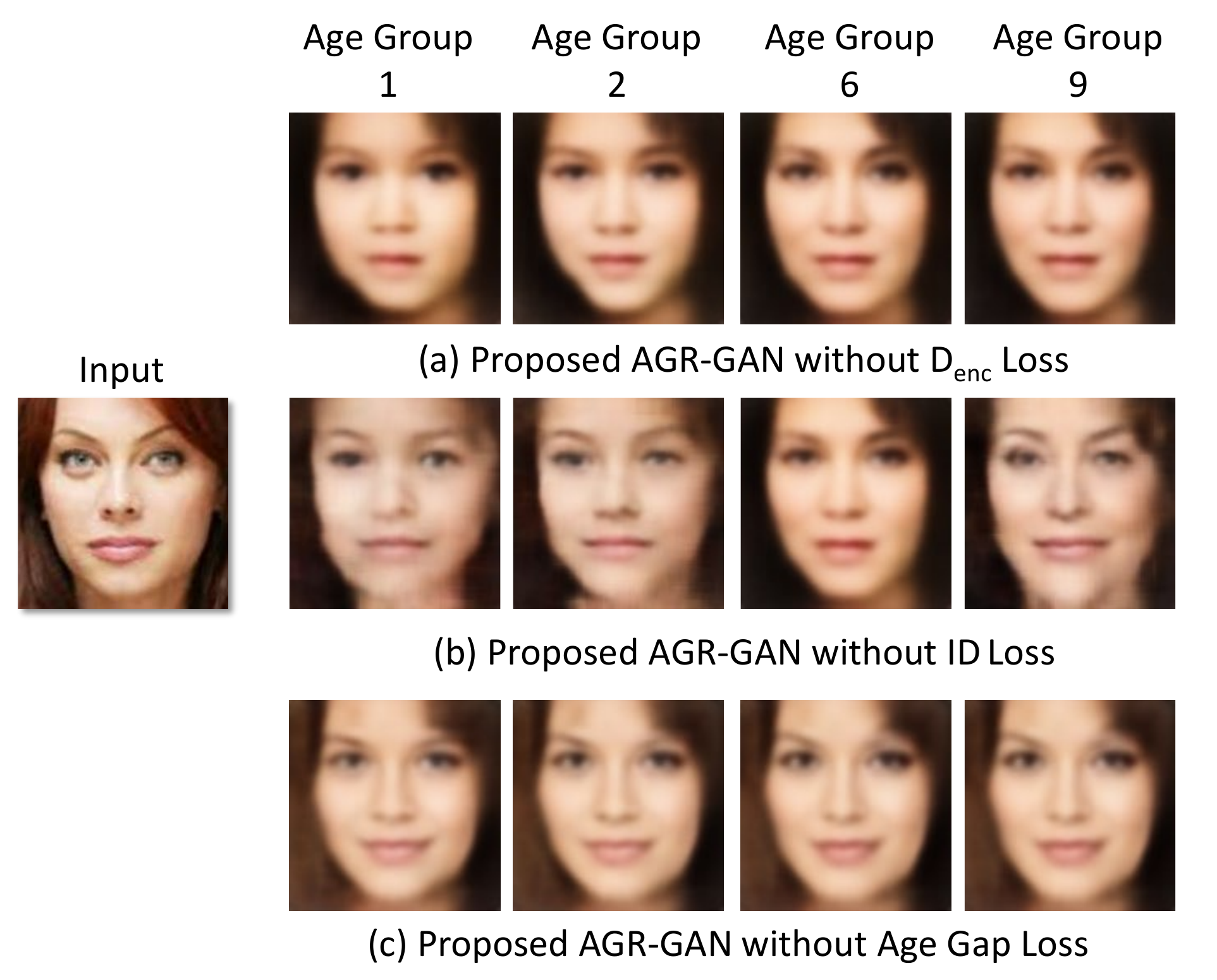}
	\caption{Illustrating the findings of ablation study to analyze the effect of removing the three loss functions in AGR-GAN. The age range of the input face image is 21-30 years). }
	\label{fig:GanSamples_Ablation_NoZ}
\end{figure}

\subsubsection{Ablation Study}
To understand the contribution of different components of the proposed AGR-GAN, an ablation study is performed. 
For this experiment, the MORPH database is utilized and the following three cases are constructed: (a) AGR-GAN without $D_{enc}$ loss, (b) AGR-GAN without identity loss, and (c) AGR-GAN without age gap loss. The performance of the proposed GAN architecture is examined using the previously described aging model evaluation and identity preservation evaluation. The results for this experiment are shown in Fig. \ref{fig:GanSamples_Ablation_NoZ} and Tables \ref{tab:ablation_agingAccu} and \ref{tab:ablation_IDAccu}.

The objective of introducing the age gap loss is to force the model to generate face images with age group as close to the target age group as possible. Therefore, when this loss term is removed from the objective function, the proposed GAN may not produce faces with the target age group. This is evident in Table \ref{tab:ablation_agingAccu}. Likewise, when the identity loss is removed from the proposed formulation, the identity preservation property of the network is removed. This may lead to lower face verification scores between the input face and generated faces. The result is shown in Table \ref{tab:ablation_IDAccu}. The $D_{enc}$ loss is a uniform prior on the latent space that ensures variations are occurring across the age groups. As is evident from Fig. \ref{fig:GanSamples_Ablation_NoZ}, removing this loss leads to images having fewer variations across the different age groups.

\begin{table}[!t]
\centering
\caption{Ablation study using aging model evaluation experiment to analyze the contribution of the age gap loss in the proposed AGR-GAN. Mean estimated age (years) of generated faces is reported. The values for column `With Age Gap Loss' are taken from Table \ref{tab:agingAccu}.}
\begin{tabular}{C{2cm}R{2.25cm}R{2.25cm}}
\toprule
\textbf{Age Group (Age Range)} & \textbf{Without Age Gap Loss} &  \textbf{With Age Gap Loss}\\ 
\midrule
1  \ \ \ (0-5)                 &  39.70      &   5.26       \\ 
2 \ (5-10)            &  39.57      &    12.18       \\ 
3  (11-15)             &  38.80      &   14.32          \\ 
4  (16-20)         &   35.42     &   17.65      \\ 
5   (21-30)        &  35.68      &   29.22      \\  
6   (31-40)               &  37.18      &   33.51       \\  
7   (41-50)               &  38.39      &   47.20        \\  
8   (51-60)               &   39.92     &   54.19         \\  
9   (61-70)               &   39.60     &    63.69         \\  
10 \ (\textgreater 70)         &   39.18     &   69.85           \\ \bottomrule
\end{tabular}
\label{tab:ablation_agingAccu}
\end{table}

\begin{table}[!t]
\centering
\caption{Ablation study using identity preservation evaluation experiment to evaluate the contribution of the identity loss in the proposed AGR-GAN. Equal error rate (\%) of face verification is reported. The values for column `With Identity Loss' are taken from Table \ref{tab:identityAccu}.}
\begin{tabular}{C{2cm}R{2.25cm}R{2.25cm}}
\toprule
\textbf{Age Group (Age Range)} & \textbf{Without Identity Loss} &  \textbf{With Identity Loss}\\ \midrule
1  \ \ \ (0-5)        &  29.30      &   12.97       \\ 
2 \ (5-10)           &  31.21      &   9.82         \\ 
3  (11-15)          &  22.19      &   8.65          \\ 
4  (16-20)          &  18.42    &    6.94     \\ 
5   (21-30)           &  19.49      &   1.63      \\ 
6   (31-40)            &  24.23      &   0.48       \\ 
7   (41-50)         &  21.63      &   1.04        \\ 
8   (51-60)        &  20.92     &   1.07         \\ 
9   (61-70)        &  26.73     &   2.38          \\ 
10 \ (\textgreater 70)         &  23.03     &   9.55           \\ \bottomrule
\end{tabular}
\label{tab:ablation_IDAccu}
\end{table}

\section{Conclusion}
In this paper, we propose a novel solution for matching face images with age gap variation. The proposed GAN based formulation involves learning a low-dimensional representation of the face. This representation is conditioned with the target age and gender labels to generate a face image. Apart from the traditional GAN adversarial loss, the training critic also involves age gap loss and identity loss between the input and the generated face images. This novel critic ensures that apart from generating photo-realistic faces, the proposed GAN reduces the age gap and preserves the identity/personality information between the input and the generated face images. Extensive experimental evaluation is performed to validate the effectiveness of the proposed AGR-GAN in matching age-separated face images on different face aging databases. The ongoing work for this research includes increasing the number and diversity of training samples in different age groups to learn accurate facial representations as the performance of such architectures depends on the amount of training data in different age groups. 

\section*{Acknowledgment}

M. Vatsa is partially supported through the Swarnajayanti Fellowship by the  Government of India.

{\small
\bibliographystyle{IEEEtran}
\bibliography{egbib}
}

\end{document}